\documentclass[letterpaper, 10 pt, conference]{ieeeconf} 
\IEEEoverridecommandlockouts
\usepackage{cite}
\usepackage{comment}
\usepackage{amsmath,amssymb,amsfonts}
\usepackage{tabularx} 
\usepackage{algorithmic}
\usepackage{graphicx}
\usepackage{textcomp}
\usepackage{xcolor}
\usepackage{multirow}
\usepackage{etoolbox}
\usepackage{balance}
\usepackage[nolist,nohyperlinks]{acronym}
\usepackage{tikz}
\usepackage{dsfont}
\usepackage{url}            
\usepackage{booktabs}       
\usepackage{nicefrac}       
\usepackage{microtype}      
\usepackage{xcolor}         
\makeatletter
\patchcmd{\@makecaption}
  {\scshape}
  {}
  {}
  {}
\makeatother

\def\BibTeX{{\rm B\kern-.05em{\sc i\kern-.025em b}\kern-.08em
    T\kern-.1667em\lower.7ex\hbox{E}\kern-.125emX}}
\usetikzlibrary{
  shapes,arrows,automata,fit,backgrounds,calc,positioning,patterns,
  decorations.pathreplacing,
  decorations.pathmorphing
}

\begin{acronym}
\acro{AUV}{autonomous underwater vehicle}
\acro{ROV}{remotely operated vehicle}
\acro{FLS}{forward-looking sonar}
\acro{SLAM}{simultaneous localization and mapping}
\acro{FoV}{field of view}
\acro{CTD}{conductivity-temperature-depth}
\acro{IMU}{inertial measurement unit}
\acro{SD}{standard deviation}
\acro{MSE}{mean squared error}
\acro{RMSE}{root mean squared error}
\acro{MAE}{mean absolute error}
\acro{NN}{neural network}
\acro{CNN}{convolutional neural network}
\acro{SSIM}{structural similarity index}
\acro{GPS}{global positioning system}
\acro{SOT}{single object tracking}
\end{acronym}

\def\-{\raisebox{.75pt}{-}}
    
\begin{document}

\setlength{\abovedisplayskip}{3pt}
\setlength{\belowdisplayskip}{3pt}

\title{\vspace*{-0mm} \LARGE \bf A Sonar-Visual Dataset for Cross-Modal Underwater Robot Perception\vspace*{-0mm}

\author{Weitung Chen$^{1}$, Phil Tinn$^{2}$, Per Gunnar Auran$^{2}$, Martin Ludvigsen$^{3}$, Peter Halland Haro$^{2}$}

\thanks{This work was supported by the Research Council of Norway under Project No.~328193.}
\thanks{$^{1}$Weitung Chen is with the Massachusetts Institute of Technology (MIT), Cambridge, MA, USA (email: \texttt{weitung@mit.edu}).}
\thanks{$^{2}$Phil Tinn, Per Gunnar Auran, and Peter Halland Haro are with SINTEF, Trondheim, Norway (email: \texttt{phil.tinn@sintef.no},  \texttt{per.gunnar.auran@sintef.no}, and \texttt{peter.haro@sintef.no}).}
\thanks{$^{3}$Martin Ludvigsen is with the Department of Marine Technology, Norwegian University of Science and Technology (NTNU), Trondheim, Norway (email: \texttt{martin.ludvigsen@ntnu.no}).}
}

\maketitle

\begin{abstract}
Underwater robots typically use both cameras and sonar for perception to leverage the rich semantic details of vision and the robust range measurements of acoustics. However, learning to map between these modalities via cross-modal prediction remains underexplored due to limited sonar-visual paired datasets. We present SOVIS, a sonar-visual dataset for cross-modal underwater perception. SOVIS comprises over 76,000 paired frames collected across 17 dives at six sites in the Trondheimfjord, supported by an end-to-end pipeline that cleans and synchronizes the cross-modal sensor data. We also introduce an interactive annotation tool designed to accelerate the labeling process for this paired data. Finally, we demonstrate a proof-of-concept cross-modal fish detection task using a small subset of labeled data, achieving a $7\times$ improvement in mAP@0.10 over a monocular camera baseline. SOVIS serves as the first step toward advancing cross-modal underwater perception research, enabling research directions such as dense sonar prediction from monocular images.
\end{abstract}

\begin{table*}[t]
\centering
\caption{Comparison of Underwater Perception Datasets. }
\label{tab:dataset_comparison}
\resizebox{\textwidth}{!}{%
\begin{tabular}{@{}llcll@{}}
\toprule
\textbf{Dataset} & \textbf{Sensing Modalities} & \textbf{Data Collection Platform} & \textbf{Environment} & \textbf{Primary Focus} \\ \midrule
SUIM \cite{islam2020suim} & Mono Cam & AUV/Diver & Ocean & Semantic Segmentation \\
NKSID \cite{nksid} & Oculus FLS & ROV & Ocean & Object Classification \\ 
DRACo-SLAM \cite{stevens_draco} & Oculus FLS + Cam & BlueROV2 & Mixed (Tank, River) & Multi-Robot SLAM \\
RGBS50 \cite{rgbs50} & Oculus FLS + Mono Cam & Fixed Platform & Pool & Single Object Tracking (SOT) \\
SOLAQUA \cite{solaqua} & Multibeam + Mono/Stereo Cam & BlueROV2 & Ocean  & Aquaculture Inspection \\ \midrule
\textbf{SOVIS (Ours)} & \textbf{Multibeam Sonar + Mono Cam} & \textbf{Blueye X3} & \textbf{Ocean} & \textbf{Cross-Modal Detection} \\ \bottomrule
\end{tabular}%
}
\vspace{-0.15in}
\end{table*}


\section{Introduction}
Underwater robots perceive the environment through two different sensing modalities: optical cameras and acoustic sonars. Cameras capture rich semantic information (color, texture, shape) that enables fine-grained recognition at close range, while multibeam sonars provide range measurements through acoustic reflections that remain robust to turbidity and low light. Neither modality alone provides a complete picture of the underwater scene. Cameras lack depth information and degrade rapidly in murky water, whereas sonars deliver sparse echo intensities without the semantic content needed for object-level understanding. If a robot could learn the mapping between these two representations, it would gain capabilities that neither sensor could provide alone. For example, the robot could perform depth estimation from a monocular image, or recognize specific objects just from acoustic sonar returns.

Such cross-modal prediction has driven major advances in terrestrial robotics, where learning to translate between LiDAR and camera views~\cite{liu2023bevfusion, li2019bevformer} has become foundational for autonomous driving. However, in the underwater world, this direction remains largely unexplored. The primary bottleneck is the lack of paired multi-modal data in underwater robotics. Unlike driving datasets that can be collected at scale on public roads, underwater data collection requires specialized robotic platforms, dive operations, and manual synchronization of sensors with fundamentally different coordinate systems. Furthermore, simulated data cannot easily close this gap: the complex physics of underwater environments, including acoustic multipath, visual scattering, and depth-dependent sound speed variations, are difficult to model accurately.
\begin{figure}[t]
\centering
\centerline{\includegraphics[width=0.485\textwidth]{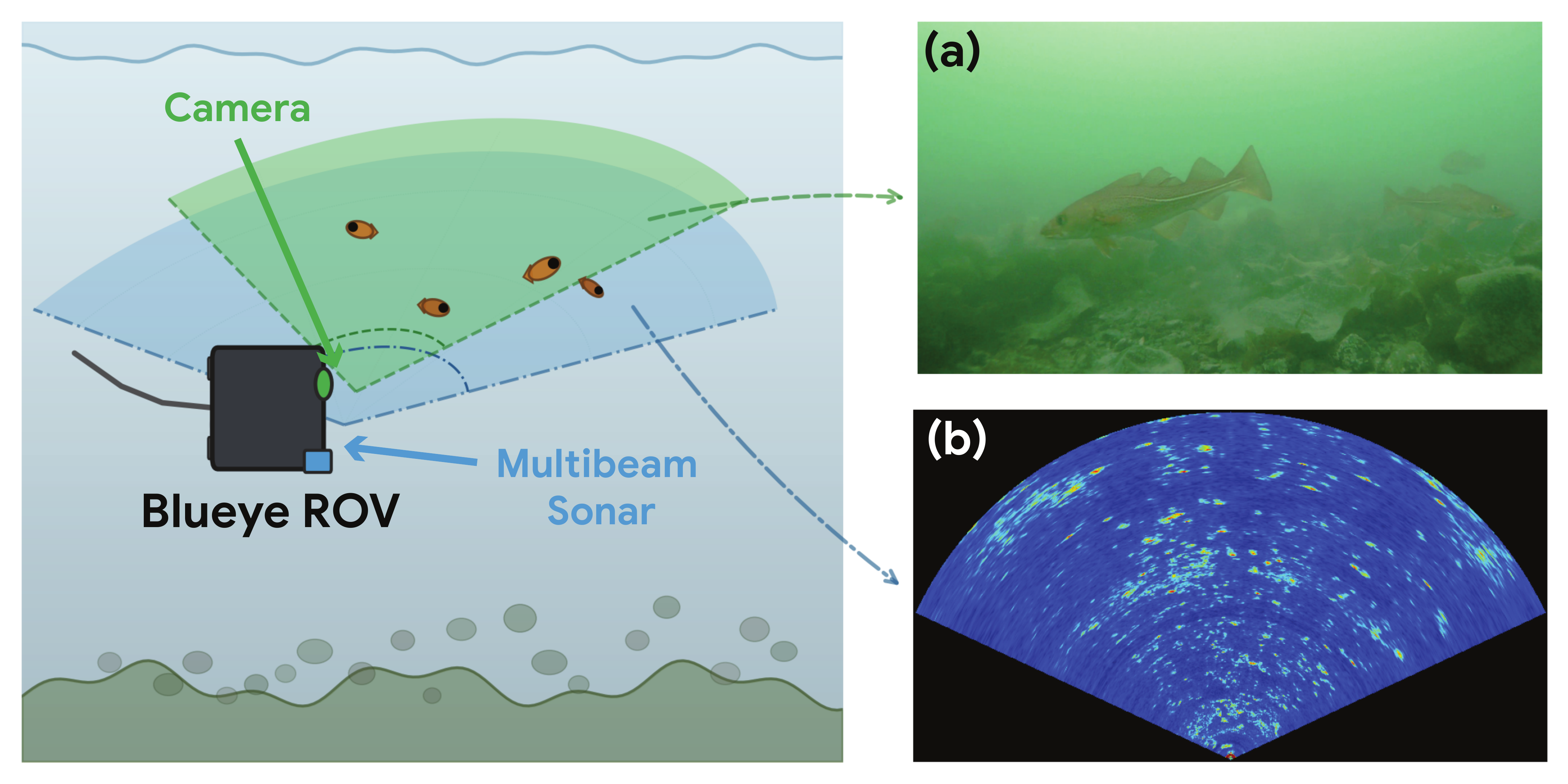}}
\vspace{-2.5mm}
\caption{\textbf{SOVIS overview.} A Blueye X3 ROV equipped with a monocular camera (green FoV) and an Oculus M750d multibeam sonar (blue FoV) collects synchronized underwater data. (a)~Camera frame showing fish right above the seafloor. (b)~Corresponding sonar return of the same scene.}
\label{fig:concept}
\vspace{-6mm}
\end{figure}

To address this challenge, we introduce \textbf{SOVIS} (SOnar-VISual), a real-world sonar-visual dataset designed for cross-modal underwater perception (Fig.~\ref{fig:concept}). We deploy a Blueye X3 ROV equipped with a monocular camera and a multibeam sonar in the Trondheimfjord, collecting over 76,000 synchronized sonar-visual pairs alongside with temperature and pressure readings. Beside the dataset, we also present a processing pipeline, an interactive cross-modal annotation tool that leverages camera-based detections to accelerate sonar labeling, and a proof-of-concept cross-modal fish detection task using a model trained on a small annotated subset.

This paper presents the following contributions:
\begin{itemize}
    \item \textbf{Sonar-Visual Dataset:} We present SOVIS, a real-world sonar-visual dataset collected in the ocean, with an end-to-end processing pipeline that cleans and synchronizes camera and sonar data. Fig.~\ref{fig:concept} shows how we collected SOVIS and an example visual-sonar pair data.  
    \item \textbf{Cross-Modal Annotation Tool:} We develop an interactive labeling interface that renders synchronized camera and sonar views side-by-side, automatically projects annotations between visual image frames and sonar frames.
    \item \textbf{Cross-Modal Fish Detection:} Using a small annotated subset produced with our tool, we demonstrate a proof-of-concept cross-modal fish detection model that achieves 46.7\% mAP@0.10, a $7\times$ improvement over a monocular camera baseline, demonstrating one cross-modal task enabled by the dataset.
\end{itemize}

\section{Related Work}
Existing underwater robot perception datasets can be separated into three categories: camera-only, acoustic-only, and multi-modal. Camera-only datasets such as SUIM~\cite{islam2020suim} provide pixel-level semantic segmentation labels for visually-guided underwater robots, while acoustic-only collections like NKSID~\cite{nksid} focus on forward-looking sonar data from ROV deployments for object classification. Multi-modal datasets pair these sensing modalities for specific downstream tasks. For example, DRACo-SLAM~\cite{stevens_draco} pairs Oculus imaging sonars with cameras on BlueROV2 platforms for multi-robot underwater SLAM, RGBS50~\cite{rgbs50} offers time-aligned sonar-camera sequences for single object tracking but was collected entirely in a controlled pool with artificial targets rather than open-water environments, and SOLAQUA~\cite{solaqua} pairs multibeam sonar with mono and stereo cameras on a BlueROV2 for aquaculture inspection. We compare these datasets with SOVIS in Table~\ref{tab:dataset_comparison}.

However, in all three categories, a critical gap persists: none of these datasets are designed for \textit{cross-modal prediction}, where models learn to map between visual and acoustic representations to allow one modality to infer the other. Existing multi-modal datasets generally treat vision and acoustics as parallel inputs for tasks like SLAM, tracking, or reconstruction, leaving cross-modal prediction largely unexplored. In terrestrial robotics and autonomous driving, cross-modal prediction has already driven major advances: methods such as Lift-Splat-Shoot~\cite{philion2020lss} and BEVFormer~\cite{li2019bevformer} produce bird's-eye-view representations from camera images alone for 3D detection and planning, while BEVFusion~\cite{liu2023bevfusion} fuses LiDAR and camera features into a shared bird's-eye-view representation that enables each modality to inform the other. Crucially, these advances were enabled by large-scale paired datasets such as nuScenes~\cite{caesar2020nuscenes} that provide dense cross-modal correspondences for training.

In the underwater domain, no analogous paired dataset with cross-modal tooling exists, leaving this research direction unexplored. SOVIS addresses this gap directly. Beyond providing synchronized sonar-visual data, SOVIS contributes a cross-modal annotation tool that projects labels between image pixels and polar beam coordinates, and a proof-of-concept fish detection benchmark. This marks the first steps toward enabling cross-modal prediction research in the underwater domain.

\section{Sonar-Visual Dataset}
\label{sec:dataset}

SOVIS is a large-scale, synchronized sonar-visual dataset consisting of time-aligned monocular camera images and multibeam sonar returns, paired with environmental measurements (water temperature and pressure). Because these parameters directly influence the speed of sound and thus acoustic range estimates, capturing them alongside the sensory feeds enables models to account for this environmental coupling. We describe the collection platform (\S\ref{sec:platform}), processing pipeline (\S\ref{sec:pipeline}), and dataset statistics (\S\ref{sec:statistics}) below.

\subsection{Data Collection Platform}
\label{sec:platform}

\begin{figure}[t]
\centering
\includegraphics[width=0.4\textwidth]{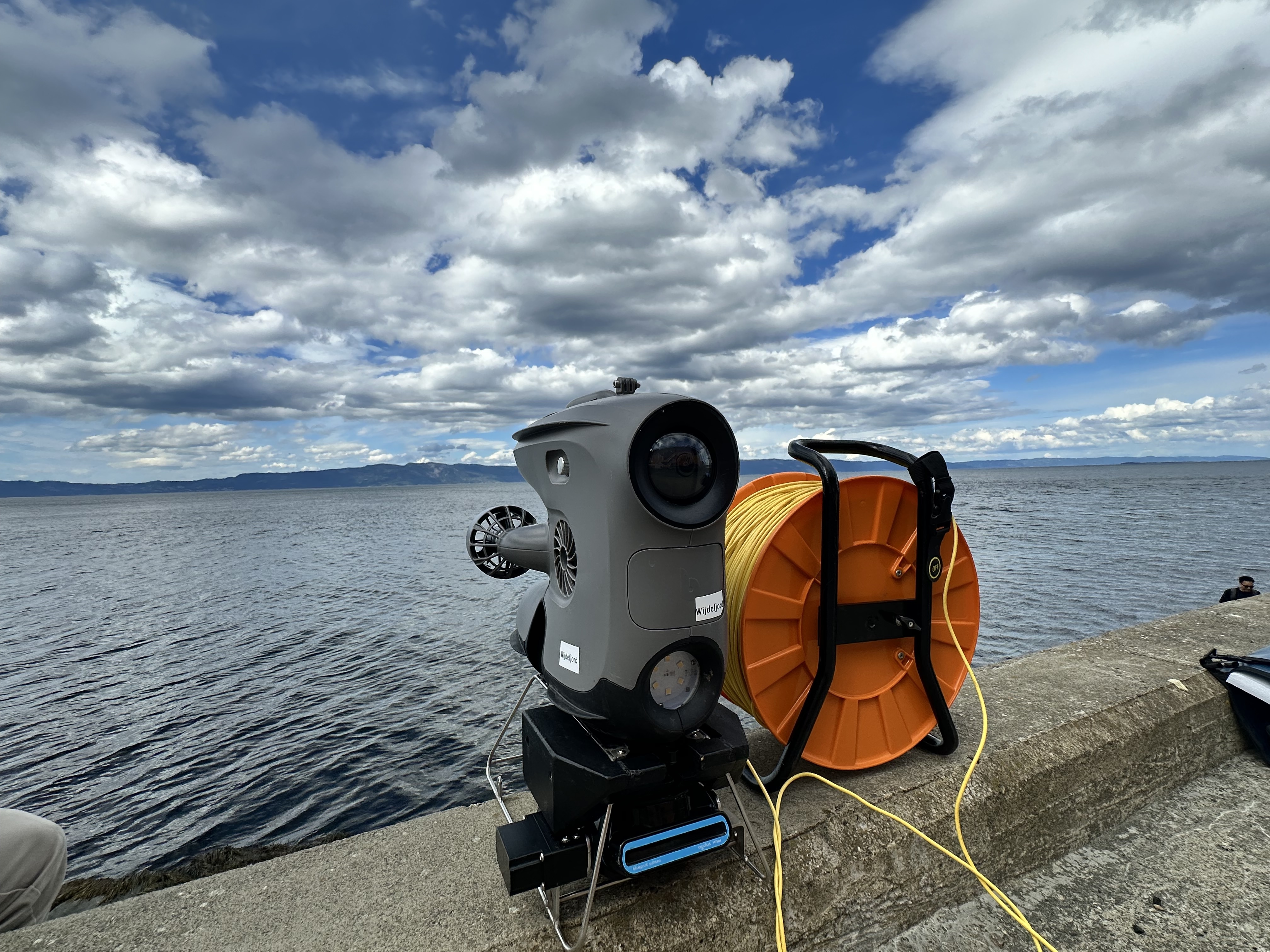}%
\vspace{-2mm}
\caption{\textbf{Data collection platform.} The Blueye X3 ROV with a tether reel. The ROV carries a monocular camera, an Oculus M750d multibeam sonar, and onboard temperature and pressure sensor. }
\label{fig:robot_photo}
\vspace{-5mm}
\end{figure}

We collect SOVIS using a Blueye X3 \ac{ROV}, a compact underwater drone designed for operations at depths up to 305\,m. The platform is equipped with three primary sensing modalities:

\textbf{Monocular Camera.} The Blueye X3 carries a built-in 1080p camera with a wide-angle lens, streaming video at 30\,fps. The camera provides the visual modality of our dataset, capturing color imagery of underwater scenes including marine life, seafloor substrate, and man-made structures.

\textbf{Multibeam Sonar.} We mount a Blueprint Subsea Oculus M750d multibeam imaging sonar on the ROV's lower payload rail. The Oculus M750d is a dual-frequency sonar; we operate in the 750\,kHz low-frequency mode, which produces fan-shaped acoustic images spanning a 130$^\circ$ horizontal aperture with 512 beams. In our deployments, we configure the sonar for a range of approximately 7\,m to maximize spatial resolution in the near field, where the camera and sonar \acp{FoV} overlap.

\textbf{Robot Onboard Sensors.} The ROV's onboard sensor suite continuously logs water temperature and hydrostatic pressure at 1\,Hz. These parameters directly govern the local speed of sound and, consequently, the accuracy of acoustic range estimates. These data are essential for physics-grounded cross-modal learning.

\begin{figure}[t]
\centering
\centerline{\includegraphics[width=0.48\textwidth]{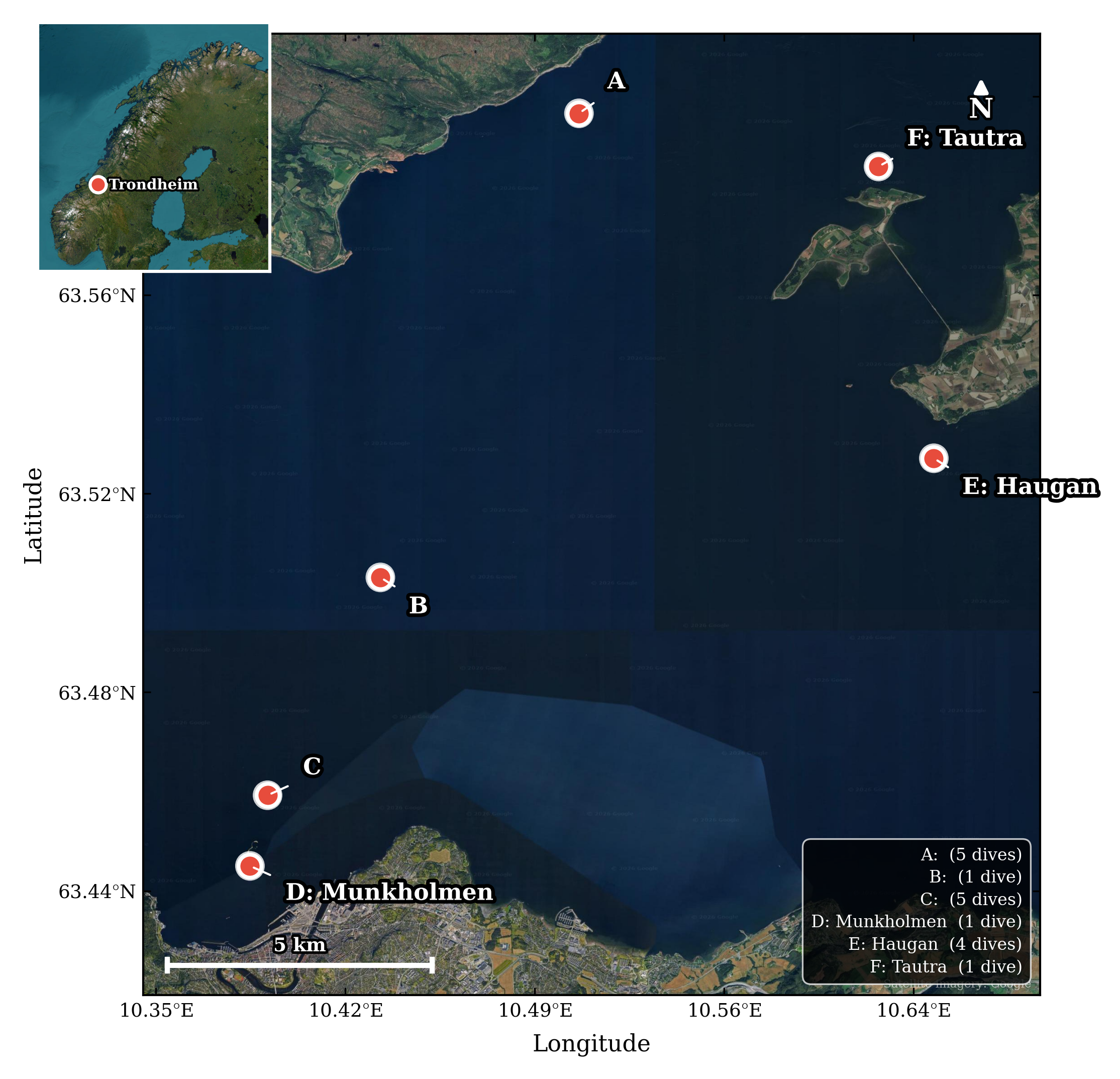}}
\vspace{-2mm}
\caption{\textbf{Deployment sites.} Satellite view of six data collection sites (A--F) in the Trondheimfjord, Norway. Site labels indicate the number of dives conducted at each location.}
\label{fig:locations}
\vspace{-5mm}
\end{figure}

\noindent We deploy the Blueye data collection platform across six sites in the Trondheimfjord near Trondheim, Norway (Fig.~\ref{fig:locations}), conducting a total of 17 dives over the course of 2 days during summer 2024. The six sites (labeled A through F) range from shallow water environment near Munkholmen island (site D) to deeper-water locations at Haugan and Tautra (sites E, F). Dive depths range from 0\,m to 91\,m, water temperatures from 7$^\circ$C to 18$^\circ$C. Fig.~\ref{fig:robot_photo} shows the Blueye X3 platform with its tethered control system at the Trondheimfjord deployment site. To support reproducibility, we document the full sensor configuration alongside the dataset, including the camera field of view and optical center, the Oculus M750d acoustic settings ($750$\,kHz, $130^\circ$ aperture, $512$ beams, ${\sim}7$\,m range at ${\sim}0.016$\,m/bin), and the per-frame temperature and pressure logs.

\subsection{Data Processing Pipeline}
\label{sec:pipeline}

Raw sensor logs from the Blueye X3 require substantial post-processing before they can serve as training data for cross-modal learning. We design an end-to-end pipeline that transforms raw multi-stream recordings into clean, synchronized sonar-visual pairs. The pipeline consists of three stages:

\textbf{1) Raw Data Ingestion.} Each dive produces three concurrent data streams: H.264-encoded video from the monocular camera, Oculus sonar files from the Oculus M750d, and dive-log telemetry from the ROV (GPS, pressure, temperature, heading). We convert the Oculus sonar files into ROS\,2 bag files to extract per-ping timestamps and beam intensity data for downstream processing.

\textbf{2) Temporal Synchronization.} We extract camera frames using OCR-based timestamp detection on the video overlay to anchor precise frame timing. Sonar pings are extracted from the ROS\,2 bags with their UTC timestamps. Because the camera and sonar originate from independent, unsynchronized clocks, we perform cross-device timestamp correction and then match each camera frame to the nearest sonar ping within a 2-second tolerance window.

\textbf{3) Sonar Image Generation.} Raw sonar pings arrive as one-dimensional arrays of $n_\text{beams} \times (n_\text{ranges} + 8)$ values in range-major layout, where the 8-byte footer per beam contains metadata. We strip the footer, reshape into a 2D polar grid of $(n_\text{ranges}, n_\text{beams})$, and crop or pad to a uniform 446 range bins ($\sim$7\,m max range at $\sim$0.016\,m/bin), producing grayscale polar sonar mask images aligned frame-by-frame with the camera stream.

\begin{figure}[t]
\centering
\includegraphics[width=0.428\textwidth]{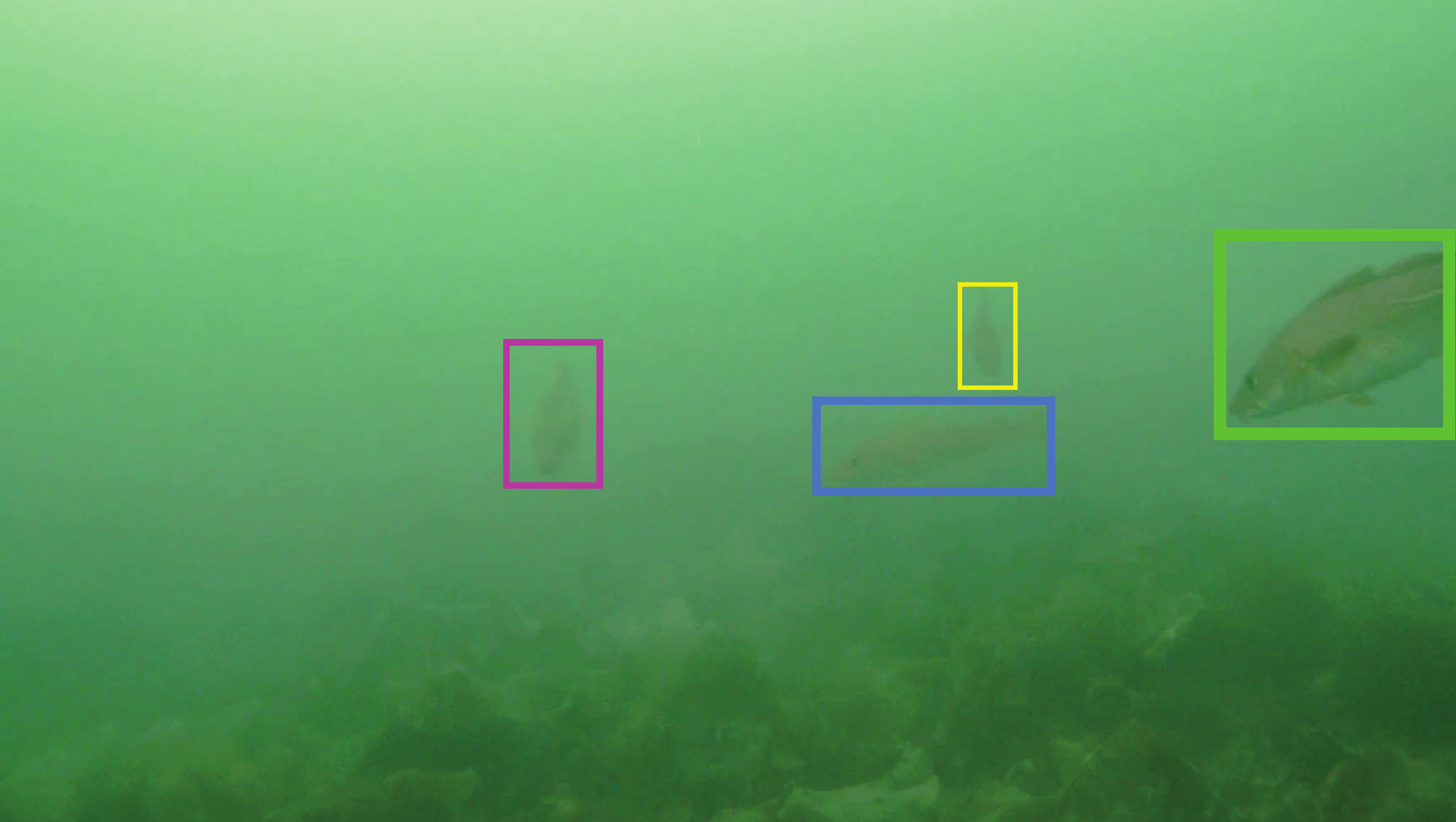} \\ \hspace{0.1mm}
\includegraphics[width=0.428\textwidth]{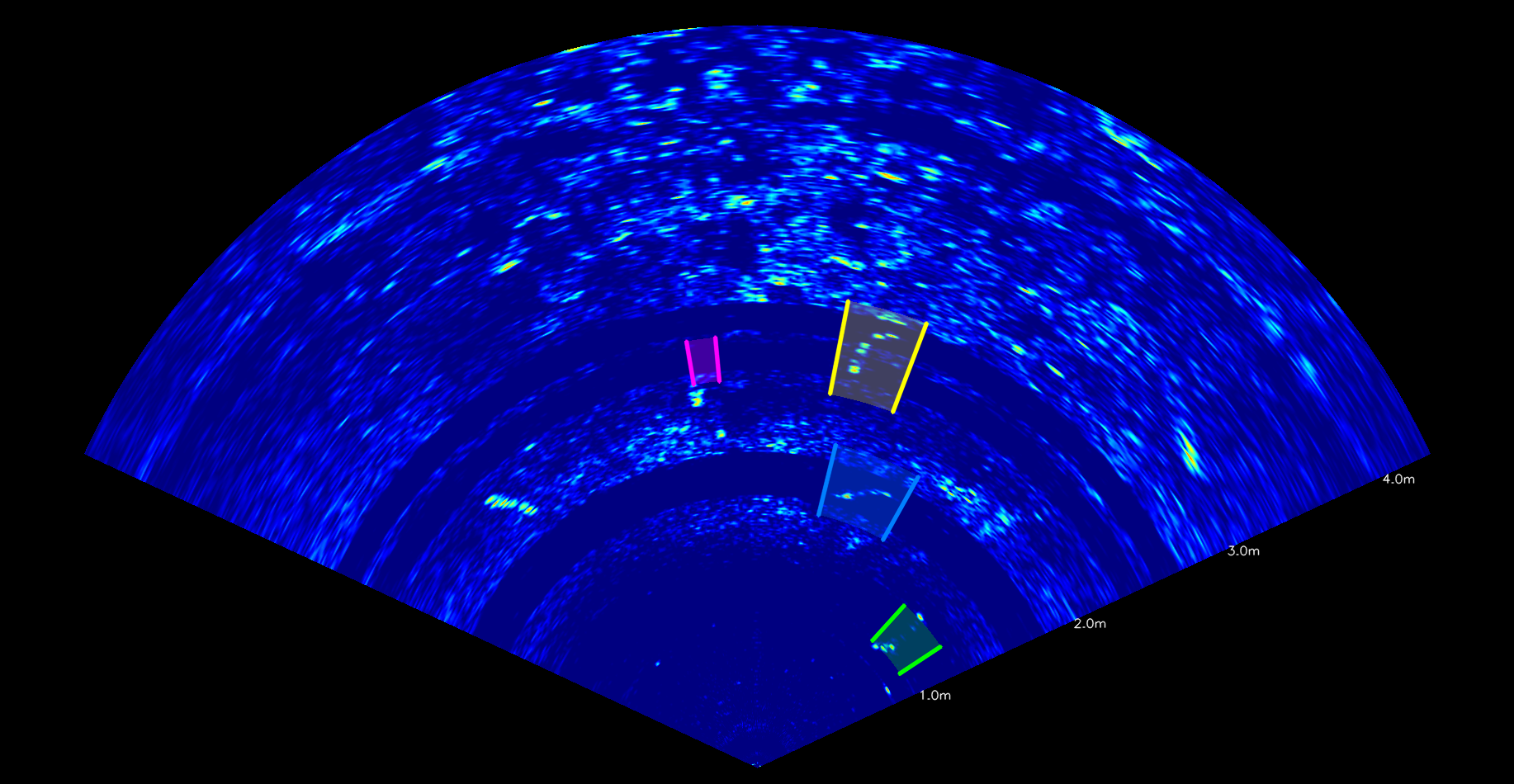}
\vspace{-2mm}
\caption{\textbf{An example sonar-visual pair.} A monocular camera frame (top) and its temporally aligned Oculus M750d sonar return (bottom), captured simultaneously during a dive. }
\label{fig:paired_example}
\end{figure}

\begin{table}[t]
\centering
\caption{Summary statistics of the SOVIS dataset.}
\label{tab:dataset_stats}
\begin{tabular}{@{}ll@{}}
\toprule
\textbf{Attribute} & \textbf{Value} \\
\midrule
Total dives & 17 \\
Deployment sites & 6 (Trondheimfjord) \\
Total recording hours & 2.1\,h \\
Synchronized pairs & $\sim$76,600 \\
Camera resolution & 1920$\times$1080 \\
Sonar operating freq. & 750\,kHz \\
Water temp. range & 7--18$^\circ$C \\
Depth range & 0--91\,m \\
\bottomrule
\end{tabular}
\vspace{-4mm}
\end{table}


\begin{figure*}[t]
\centering
\includegraphics[width=0.95\textwidth]{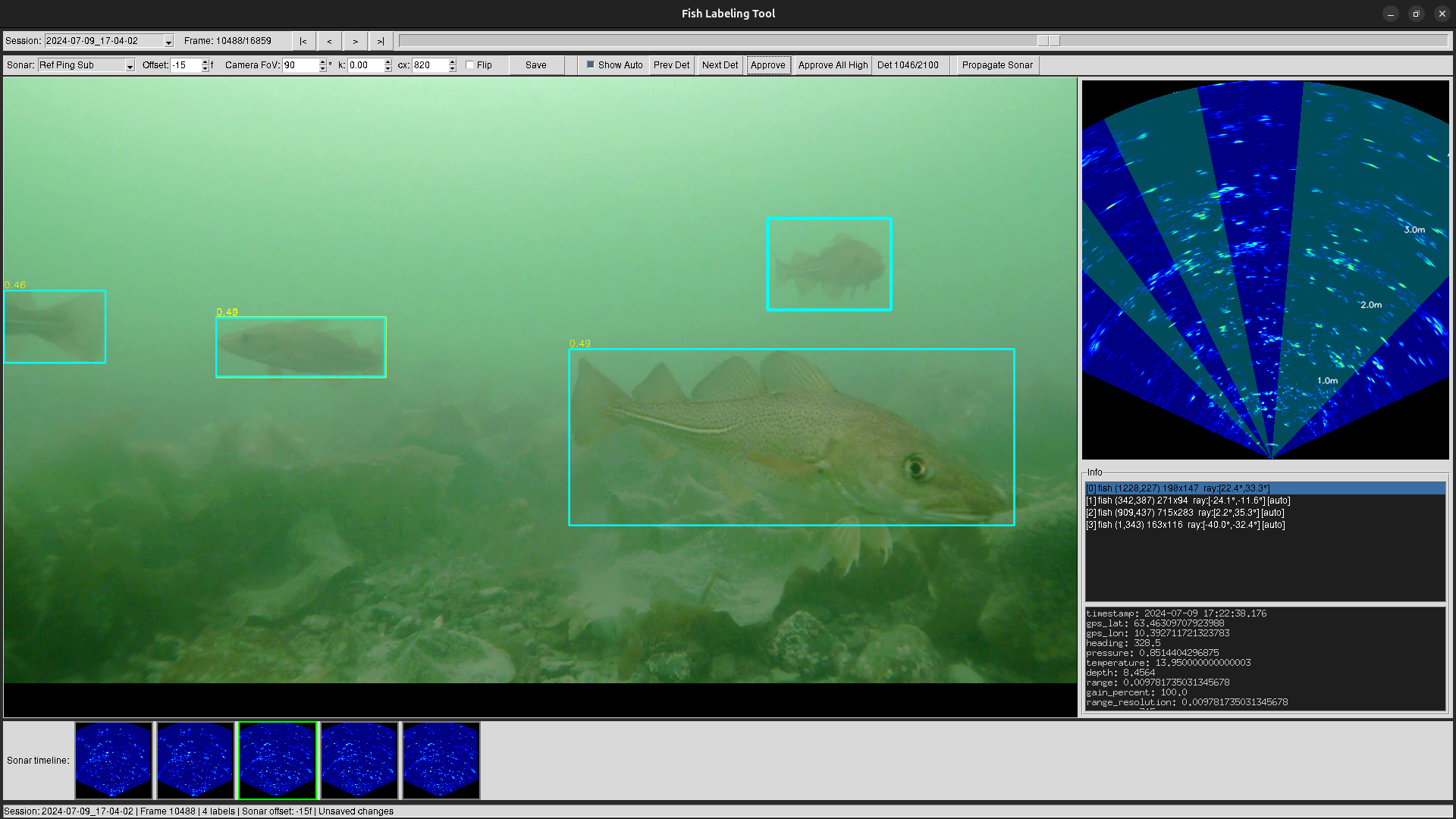}
\vspace{-2mm}
\caption{\textbf{Cross-modal annotation tool.} The interface displays the camera frame (left) with fish bounding boxes in cyan and the sonar fan visualization (top right) with angular bands highlighting detected regions. The bottom strip shows $\pm$2 neighboring sonar frames for temporal context. When an annotator draws a bounding box on the camera image, the tool automatically projects it to the corresponding angular extent in the sonar view using a parameterized lens model. Auto-detected labels are color-coded by confidence (green: high, yellow: medium, red: low) for rapid review.}
\label{fig:label_tool}
\vspace{-4mm}
\end{figure*}

\subsection{Dataset Structure and Statistics}
\label{sec:statistics}

After processing, SOVIS contains 2.1 hours of strictly synchronized sonar-visual data across all 17 dives. Table~\ref{tab:dataset_stats} summarizes the key statistics. The dataset comprises approximately 76,600 synchronized frame-beam pairs spanning depth ranges from 0\,m to 91\,m. 

Fig.~\ref{fig:paired_example} shows a representative synchronized pair from the dataset. The camera image captures several fish swimming above the seafloor. The corresponding sonar image, captured at the same instant, reveals the acoustic reflectance pattern of the scene, illustrating the complementary information provided by the two modalities: the camera provides rich visual detail while the sonar resolves range and detects targets beyond visual range. The bounding boxes in the figures are color coded (pink, blue, yellow, and green) to show the corresponding fish acoustic returns on the sonar image. 


\section{Cross-Modal Annotation Tool}
\label{sec:annotation}

A key challenge in building cross-modal datasets is annotation: camera images and sonar returns use fundamentally different coordinate systems (dense Cartesian pixels vs.\ sparse polar beams), making it difficult to label corresponding objects across modalities. To address this, we develop an interactive annotation tool that enables efficient cross-modal labeling (Fig.~\ref{fig:label_tool}).

\subsection{Interface Design}
The tool presents synchronized camera and sonar views side-by-side (Fig.~\ref{fig:label_tool}). The left panel displays the camera frame with overlaid bounding boxes; the right panel renders the sonar data as a fan-shaped polar image via a precomputed polar-to-Cartesian remap table, with distance rings annotated in meters. A temporal thumbnail strip below the sonar panel shows $\pm$2 neighboring sonar frames to aid in distinguishing moving targets (e.g., fish) from static background.

\subsection{Camera-to-Sonar Projection}

When an annotator draws a bounding box on the camera image, the tool automatically computes the corresponding angular location in the sonar's polar coordinate system. We model this projection using a parameterized lens model. Given a pixel at horizontal position $x$ in the camera image with optical center $c_x$, the normalized coordinate $x_n = (x - c_x)/c_x$ is mapped to an angular coordinate via:
\begin{equation}
    \theta = \frac{\text{FoV}}{2} \cdot x_n \cdot \left(1 + k\,(1 - x_n^2)\right)
\end{equation}
where FoV is the camera's horizontal field of view (default $90^\circ$) and $k \in [-0.5, 0.5]$ is a distortion parameter.\footnote{$k{=}0$: equidistant projection; $k{>}0$: rectilinear-like; $k{<}0$: barrel distortion.} The parameters FoV, $k$, and $c_x$ are adjustable in real time via the tool's interface, allowing the operator to tune the projection to match the observed camera-sonar correspondence. The projected angles are clamped to the sonar's $\pm 65^\circ$ aperture.

\begin{figure*}[t]
\centering
\vspace{-3mm}
\includegraphics[width=0.99\textwidth]{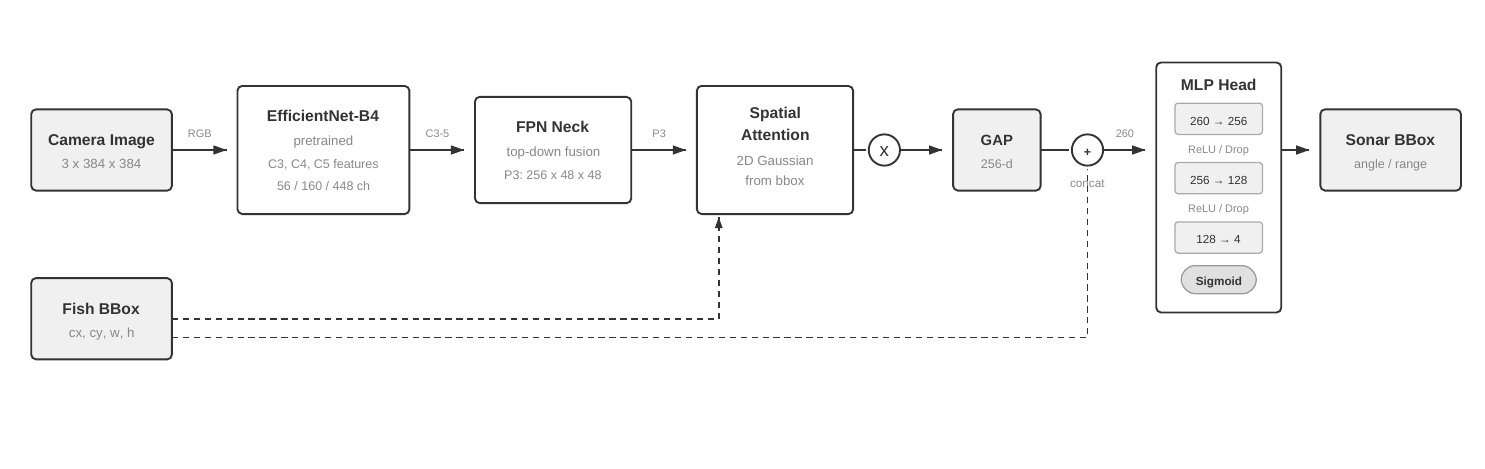}
\vspace{-13mm}
\caption{\textbf{SOVISFishNet architecture.} An EfficientNet-B4 encoder extracts multi-scale features, which are fused by an FPN neck into a unified feature map. A 2D Gaussian spatial attention mask, generated from the camera fish bounding box, gates the feature map. After global average pooling (GAP), the visual features are concatenated with the normalized bounding box coordinates and decoded by an MLP head into four sonar bbox parameters.}
\label{fig:architecture}
\vspace{-5mm}
\end{figure*}

\subsection{Semi-Auto Labeling}

To reduce manual annotation effort, we integrate a three-stage auto-detection pipeline:

\textbf{Stage~1 (Camera Detection):} A YOLOv8m object detector~\cite{yolov8} identifies candidate fish in camera frames at a low confidence threshold to maximize recall.

\textbf{Stage~2 (Sonar Motion Detection):} We detect moving acoustic targets via temporal median subtraction on the sonar mask, followed by adaptive thresholding and connected-component extraction. Blobs are filtered by size and temporal consistency across consecutive frames.

\textbf{Stage~3 (Cross-Modal Fusion):} Camera and sonar detections are matched by angular overlap, producing a fused confidence score that weights both the visual detector confidence and sonar intensity. A temporal IoU tracker links detections across frames, boosting confidence for persistent tracks.

The tool displays auto-detections color-coded by confidence tier, allowing an operator to rapidly approve, reject, or refine annotations. In practice, this semi-automatic workflow reduces labeling time by approximately an order of magnitude compared to fully manual annotation.

\section{Cross-Modal Fish Detection}
\label{sec:detection}
To demonstrate how SOVIS enables cross-modal robotic perception underwater, we present a proof-of-concept cross-modal fish detection task. Using the semi-automatic annotation tool described in Section~\ref{sec:annotation}, we efficiently scan SOVIS's 76,000+ frames to identify fish that exhibit both visual and acoustic signatures, and label a small subset of these instances for supervised training. Given a fish detection in the camera frame, the objective is to predict where the corresponding acoustic signature appears in the sonar's polar coordinate system.

\subsection{Task Formulation}

We select all frames containing a confirmed fish annotation (camera bounding box matched with a sonar bounding box, confidence $\geq$0.5), yielding 306 labeled instances across 256 frames from two dive sessions. We split these into 276 training and 30 validation instances. For each instance, the model receives a camera image and a fish bounding box, and must predict the corresponding sonar bounding box (angle center, angle width, range center, range width). We frame this as camera-to-sonar fish localization, in which the sonar bounding boxes provide supervision rather than model input.

\subsection{Model Architecture}

We propose SOVISFishNet, a model that predicts sonar bounding boxes directly from camera images and fish bounding box locations. As illustrated in Fig.~\ref{fig:architecture}, SOVISFishNet employs an ImageNet-pretrained EfficientNet-B4~\cite{tan2019efficientnet} backbone to extract multi-scale visual features, which are fused into a unified feature map by a Feature Pyramid Network (FPN) neck through top-down lateral connections.

To focus the model on the relevant image region, we generate a 2D Gaussian spatial attention mask from the camera bounding box and apply it to gate the feature map. The attended features are then reduced via global average pooling, concatenated with the normalized bounding box coordinates, and decoded by an MLP head into four sonar bounding box parameters.\footnote{SOVISFishNet has 18.2M parameters. We train with smooth-$L_1$ loss, AdamW optimizer (backbone lr $2{\times}10^{-5}$, decoder lr $2{\times}10^{-4}$), OneCycleLR scheduling, and encoder freezing for the first 5 of 120 total epochs.}

\section{Results}
\label{sec:results}

We evaluate SOVISFishNet on the 30-instance validation set using mean average precision at IoU threshold 0.10 (mAP@0.10), along with median angle and range errors. We use the lower IoU threshold because sonar bounding boxes are small relative to the full $130^\circ \times 7$\,m sonar field of view, so even small localization errors significantly reduce IoU.

\subsection{Quantitative Evaluation}

We compare SOVISFishNet against a \textit{geometric baseline} that maps camera detections to sonar coordinates using only the known sensor geometry. The baseline computes the sonar angle by projecting the camera bounding box center through the horizontal field-of-view mapping (pixel $x$ $\to$ sonar angle via $90^\circ$ HFOV), and sets the range to the training-set mean (1.80\,m) for all predictions since monocular images provide no depth cue.

\begin{table}[t]
\centering
\caption{\textbf{Cross-modal fish detection results.} SOVISFishNet vs.\ the geometric baseline on sonar-domain fish bounding box location prediction.}
\label{tab:results}
\begin{tabular}{@{}lccc@{}}
\toprule
\textbf{Method} & \textbf{mAP@.10} & \textbf{Angle Err} & \textbf{Range Err} \\
\midrule
Geometric Baseline & 0.067 & $\mathbf{3.3^\circ}$ & 0.644\,m \\
\textbf{Ours (SOVISFishNet)} & \textbf{0.467} & $4.8^\circ$ & \textbf{0.102\,m} \\
\bottomrule
\end{tabular}
\vspace{-4mm}
\end{table}

Table~\ref{tab:results} reports the results. The geometric baseline achieves 6.7\% mAP@0.10 with a median angle error of $3.3^\circ$ and range error of 0.644\,m. Its angle prediction is reasonable because it directly applies the known camera-to-sonar angular mapping, but it cannot resolve range and simply defaults to the training-set mean.

SOVISFishNet achieves 46.7\% mAP@0.10, a $7\times$ improvement over the geometric baseline. The model predicts range $6.3\times$ more accurately (0.102\,m vs.\ 0.644\,m median error) by learning visual depth cues from image content such as apparent fish size and scene context. The geometric baseline achieves slightly better angle prediction ($3.3^\circ$ vs.\ $4.8^\circ$), which is expected given that it directly uses the calibrated angular mapping while our model must learn this relationship from only 276 training samples. These results demonstrate that even with a small labeled subset, cross-modal learning on SOVIS can recover meaningful camera-to-sonar correspondences, illustrating one cross-modal task that SOVIS and our annotation tool enable. The full synchronized dataset remains available for broader tasks such as dense pixel-level sonar prediction.

\subsection{Qualitative Results}

\begin{figure}[t]
\centering
\includegraphics[width=0.485\textwidth]{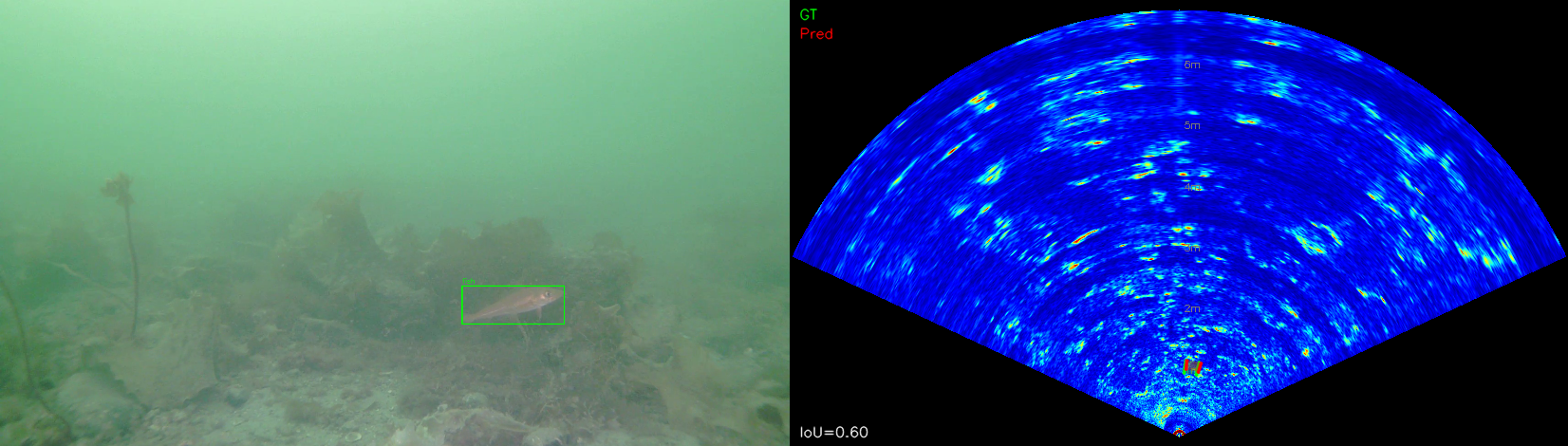}\\[1.5pt]
\includegraphics[width=0.485\textwidth]{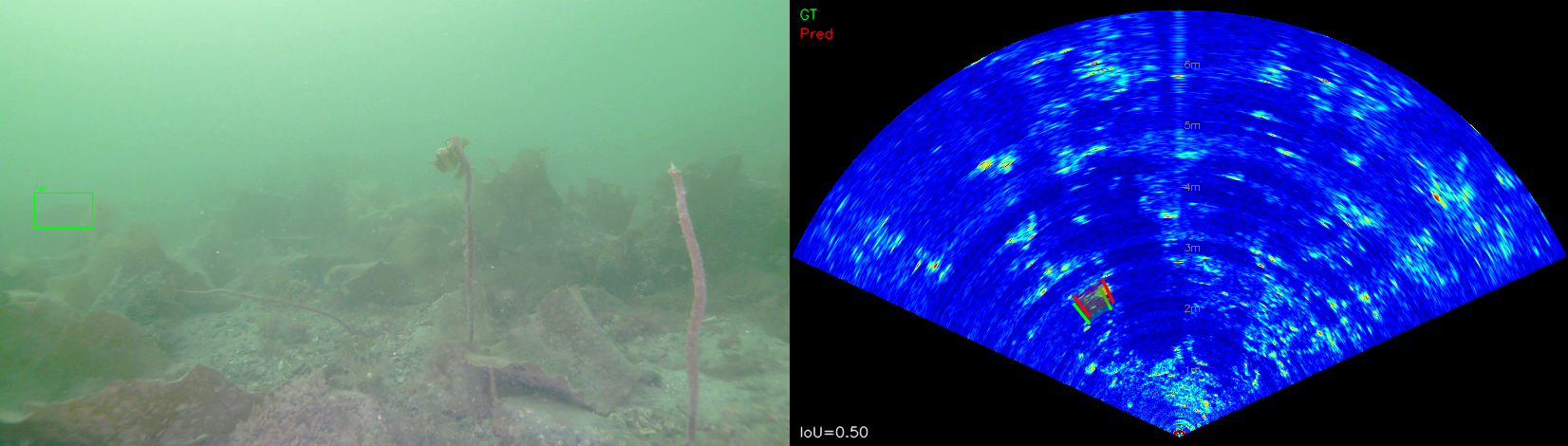}
\vspace{-2mm}
\caption{\textbf{Qualitative cross-modal detection results.} Each panel shows the camera frame (left) with detected fish (green box) and the sonar fan view (right) with ground-truth (green) and predicted (red) bounding boxes. The model correctly localizes the acoustic signature of the visually detected fish, achieving IoU of 0.60 (top) and 0.50 (bottom).}
\label{fig:qualitative}
\vspace{-5mm}
\end{figure}

Fig.~\ref{fig:qualitative} shows two representative cross-modal detection results. In both cases, the model correctly localizes the acoustic signature of the visually detected fish in the sonar domain, producing predicted bounding boxes that overlap well with the ground truth (IoU 0.60 and 0.50). The spatial attention mechanism, guided by the camera bounding box, effectively focuses the model on the relevant image region, while the learned features resolve the range dimension that the geometric baseline cannot capture.

\section{Conclusion}
\label{sec:conclusion}

We presented SOVIS, a real-world sonar-visual dataset for cross-modal underwater robot perception. Alongside the dataset, we introduced an interactive cross-modal annotation tool that leverages camera-based visual detection to accelerate sonar annotation, and demonstrated its utility by labeling 306 fish instances for a proof-of-concept cross-modal detection task. Even with this small labeled subset, our model achieves 46.7\% mAP@0.10, a $7\times$ improvement over a monocular camera baseline, validating that meaningful camera-to-sonar correspondences can be learned from SOVIS. While this work focuses on bounding-box-level fish detection as an initial demonstration, the full synchronized dataset enables a broader set of cross-modal tasks. In particular, the dense pixel-level alignment between camera frames and sonar masks provides a foundation for training models that predict sonar intensity maps directly from monocular images, enabling camera-only robots to infer acoustic scene structure without an active sonar. We believe SOVIS, together with its annotation tooling and a fish detection benchmark, represents a step toward advancing underwater robotic perception through cross-modal prediction.

\balance
\bibliographystyle{IEEEtran}
\bibliography{reference.bib}

\end{document}